\documentclass[11pt,a4paper]{article}
\usepackage[hyperref]{emnlp2020}
\usepackage{times}
\usepackage{latexsym}

\usepackage{pifont}
\usepackage[normalem]{ulem}
\usepackage{microtype}
\usepackage{enumitem}
\usepackage{booktabs}
\usepackage{multirow}
\usepackage{multicol}
\usepackage{graphicx}
\usepackage{array}
\usepackage{pgfplots}
\usepackage{amsmath}
\usepackage{algorithm}
\usepackage{url}
\usepackage{makecell}
\usepackage{subcaption}
\usepackage{eurosym}
\usepackage{gensymb}
\usepackage{amsmath}
\usepackage{amsfonts}
\usepackage{amsthm,amssymb,amsopn,bm}
\usepackage[noend]{algpseudocode}

\usepackage{xcolor,colortbl}
\usepackage{soul}
\usepackage{framed}

\makeatletter
\newcommand\footnoteref[1]{\protected@xdef\@thefnmark{\ref{#1}}\@footnotemark}
\makeatother

\newcommand{\mb}[1]{\boldsymbol{\mathbf{#1}}}
\newcommand{\loss}{\ensuremath\mathcal{L}}

\DeclareMathOperator*{\argmin}{arg\,min}

\definecolor{adversarial}{rgb}{0.90, 0.02, 0.03}
\newcommand{\PreserveBackslash}[1]{\let\temp=\\#1\let\\=\temp}
\newcolumntype{C}[1]{>{\PreserveBackslash\centering}p{#1}}
\newcolumntype{R}[1]{>{\PreserveBackslash\raggedleft}p{#1}}
\newcolumntype{L}[1]{>{\PreserveBackslash\raggedright}p{#1}}

\makeatletter
\def\blfootnote{\xdef\@thefnmark{}\@footnotetext}
\makeatother

\aclfinalcopy

\definecolor{domain}{HTML}{FFC7BF}
\definecolor{date}{HTML}{FFE9BF}
\definecolor{authors}{HTML}{AFFFCC}
\definecolor{headline}{HTML}{C0DEFF}
\definecolor{body}{HTML}{E3C0FF}

\makeatletter
 \def\SOUL@hlpreamble{%
 \setul{}{2.4ex}%
 \let\SOUL@stcolor\SOUL@hlcolor
 \SOUL@stpreamble
 }
\makeatother

\newcommand{\hlc}[2][yellow]{{%
    \colorlet{foo}{#1}%
    \sethlcolor{foo}\hl{#2}}%
}

\newcommand{\google}[1]{\hlc[domain]{#1}}

\newif\ifcomments
\ifcomments
    \providecommand{\eric}[1]{{\protect\color{magenta}{[Eric: #1]}}}
    \providecommand{\mitchell}[1]{{\protect\color{purple}{\bf [Mitchell: #1]}}}
\else
    \providecommand{\eric}[1]{}
    \providecommand{\mitchell}[1]{}
\fi
\title{\mbox{Imitation Attacks and Defenses for Black-box Machine Translation Systems}}
\author{\makecell{Eric Wallace \hspace{0.5cm}  Mitchell Stern \hspace{0.5cm} Dawn Song}\\
UC Berkeley\\
\{\href{mailto:ericwallace@berkeley.edu}{\tt ericwallace},
\href{mailto:mitchell@berkeley.edu}{\tt mitchell},
\href{mailto:dawnsong@berkeley.edu}{\tt dawnsong}\}\href{ericwallace@berkeley.edu}{\tt @berkeley.edu}}

\date{}

\begin{document}
\maketitle
\begin{abstract}
Adversaries may look to \emph{steal} or \emph{attack} black-box NLP systems, either for financial gain or to exploit model errors. One setting of particular interest is machine translation (MT), where models have high commercial value and errors can be costly.
We investigate possible exploits of black-box MT systems and explore a preliminary defense against such threats.
We first show that MT systems can be stolen by querying them with monolingual sentences and training models to imitate their outputs. Using simulated experiments, we demonstrate that MT model stealing is possible even when imitation models have different input data or architectures than their target models. Applying these ideas, we train imitation models that reach within 0.6 BLEU of three production MT systems on both high-resource and low-resource language pairs. We then leverage the similarity of our imitation models to transfer adversarial examples to the production systems. We use gradient-based attacks that expose inputs which lead to semantically-incorrect translations, dropped content, and vulgar model outputs. To mitigate these vulnerabilities, we propose a defense that modifies translation outputs in order to misdirect the optimization of imitation models. This defense degrades the adversary's BLEU score and attack success rate at some cost in the defender's BLEU and inference speed.
\end{abstract}

\section{Introduction}

\begin{figure*}[tbh]
\centering
\includegraphics[trim={0cm 0.3cm 0cm 0.3cm},clip, width=0.88\textwidth]{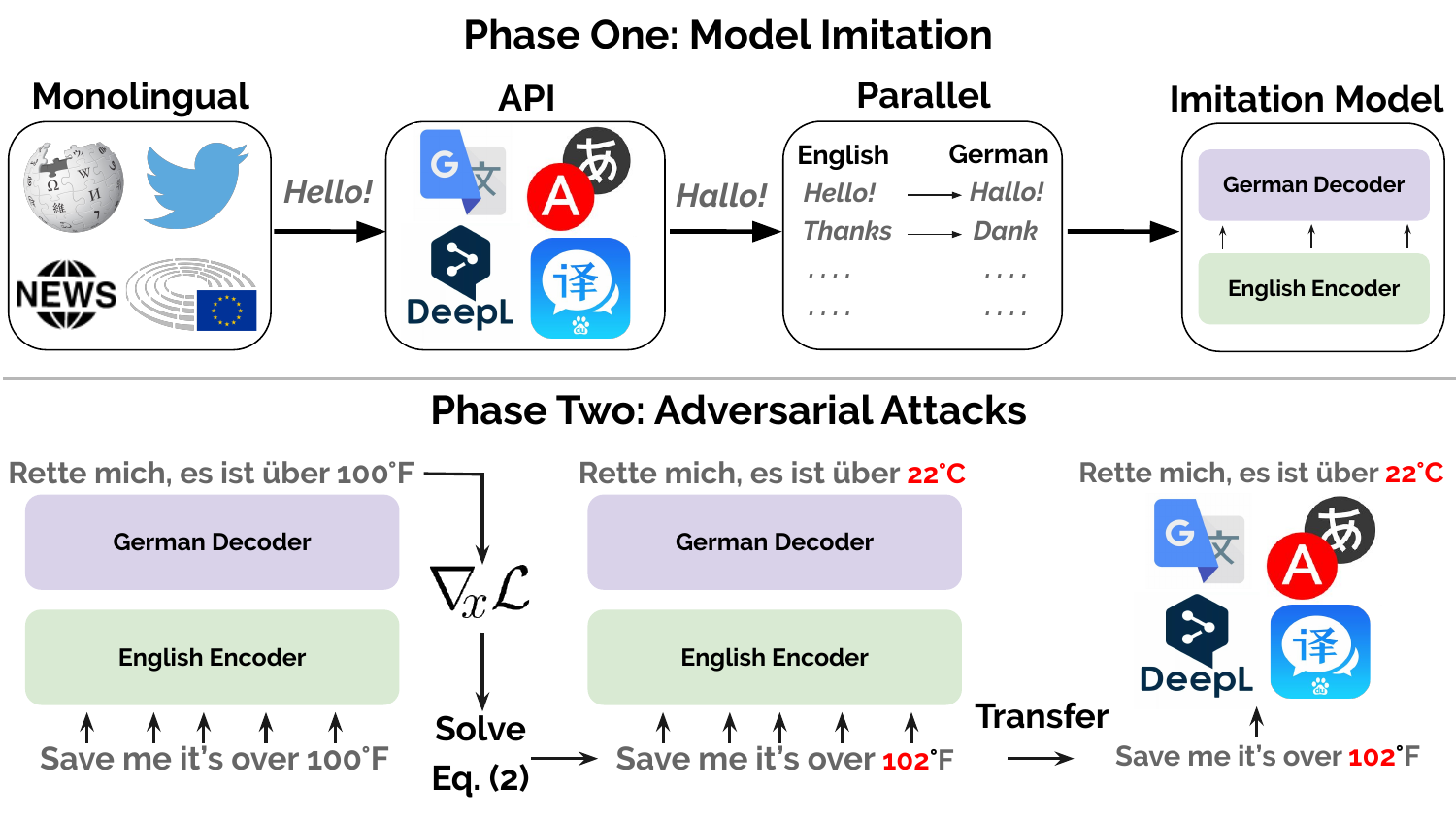}
\vspace{-0.25cm}
\caption{\emph{Imitating and attacking an English$\to$German MT system.} In phase one (model imitation), we first select sentences from English corpora (e.g., Wikipedia), label them using the black-box API, and then train an imitation model on the resulting data.
In phase two (adversarial attacks), we generate adversarial examples against our imitation model and transfer them to the production systems. For example, we find an input  perturbation that causes Google to produce a factually incorrect translation (all attacks work as of April 2020).}
\label{fig:overview}
\end{figure*}

NLP models deployed through APIs (e.g., Google Translate) can be lucrative assets for an organization. These models are typically the result of a considerable investment---up to millions of dollars---into private data annotation and algorithmic improvements.
Consequently, such models are kept hidden behind black-box APIs to protect system integrity and intellectual property.

We consider an adversary looking to \emph{steal} or \emph{attack} a black-box NLP system. Stealing a production model allows an adversary to avoid long-term API costs or launch a competitor service. Moreover, attacking a system using adversarial examples~\cite{szegedy2013intriguing} allows an adversary to cause targeted errors for a model, e.g., bypassing fake news filters or causing systems to output malicious content that may offend users and reflect poorly on system providers. In this work, we investigate these two exploits for black-box machine translation (MT) systems: we first steal (we use ``steal'' following \citealt{tramer2016stealing}) production MT systems by training \textit{imitation models}, and we then use these imitation models to generate adversarial examples for production MT systems.

We create imitation models by borrowing ideas from knowledge distillation~\cite{hinton2015distilling}: we query production MT systems with monolingual sentences and train imitation (i.e., student) models to mimic the system outputs (top of Figure~\ref{fig:overview}). We first experiment with simulated studies which demonstrate that MT models are easy to imitate (Section~\ref{sec:simulated}). For example, imitation models closely replicate the target model outputs even when they are trained using different architectures or on out-of-domain queries. Applying these ideas, we imitate production systems from \texttt{Google}, \texttt{Bing}, and \texttt{Systran} with high fidelity on English$\to$German and Nepali$\to$English. For example, \texttt{Bing} achieves 32.9 BLEU on WMT14 English$\to$German and our imitation achieves 32.4 BLEU.

We then demonstrate that our imitation models aid adversarial attacks against production MT systems (Section~\ref{sec:adversarial}). In particular, the similarity of our imitation models to the production systems allows for direct transfer of adversarial examples obtained via gradient-based attacks. We find small perturbations that cause targeted mistranslations (e.g., bottom of Figure~\ref{fig:overview}), nonsense inputs that produce malicious outputs, and universal phrases that cause mistranslations or dropped content.

The reason we identify vulnerabilities in NLP systems is to robustly patch them. To take steps towards this, we create a defense which finds alternate translations that cause the optimization of the imitation model to proceed in the wrong direction (Section~\ref{sec:defenses}). These alternate translations degrade the imitation model's BLEU score and the transfer rate of adversarial examples at some cost in the defender's BLEU and inference speed.
\section{How We Imitate MT Models}\label{sec:method}
We have query access to the predictions (but no probabilities or logits) from a \emph{victim} MT model. This victim is a black box: we are unaware of its internals, e.g., the model architecture, hyperparameters, or training data. Our goal is to train an \emph{imitation model}~\cite{orekondy2019knockoff} that achieves comparable accuracy to this victim on held-out data. Moreover, to enhance the transferability of adversarial examples, the imitation model should be functionally similar to the victim, i.e., similar inputs translate to similar outputs.\smallskip

\noindent \textbf{Past Work on Distillation and Stealing} This problem setup is closely related to model \textit{distillation}~\cite{hinton2015distilling}: training a student model to imitate the predictions of a teacher. Distillation has widespread use in MT, including reducing architecture size~\cite{kim2016sequence,kim2019research}, creating multilingual models~\cite{tan2018multilingual}, and improving non-autoregressive generation~\cite{ghazvininejad2019constant,stern2019insertion}. Model stealing differs from distillation because the victim's (i.e., teacher's) training data is unknown. This causes queries to typically be out-of-domain for the victim. Moreover, because the victim's output probabilities are unavailable for most APIs, imitation models cannot be trained using distribution matching losses such as KL divergence, as is common in distillation.\looseness=-1

Despite these challenges, prior work shows that model stealing is possible for simple classification~\cite{lowd2005stealing,tramer2016stealing}, vision~\cite{orekondy2019knockoff}, and language tasks~\cite{krishna2019thieves,pal2019framework}. In particular, \citet{pal2019framework} steal text classifiers and \citet{krishna2019thieves} steal reading comprehension and textual entailment models; we extend these results to MT and investigate how model stealing works for production systems.\smallskip

\noindent \textbf{Our Approach} We assume access to a corpus of monolingual sentences. We select sentences from this corpus, query the victim on each sentence, and obtain the associated translations. We then train an imitation model on this ``labeled'' data.
\section{Imitating Black-box MT Systems}\label{sec:simulated}

We first study imitation models through simulated experiments: we train victim models, query them as if they are black boxes, and then train imitation models to mimic their outputs. In Section~\ref{sec:real}, we turn to imitating production systems.

\subsection{Research Questions and Experiments}

In practice, the adversary will not know the victim's model architecture or source data. We study the effect of this with the following experiments:
\begin{itemize}[nosep,leftmargin=2mm]
    \item We use the same architecture, hyperparameters, and source data as the victim (\emph{All Same}).
    \item We use the same architecture and hyperparameters as the victim, but use an out-of-domain (OOD) source dataset (\emph{Data Different}).
    \item We use the same source data but a different architecture, either (1) the victim is a Transformer and the imitator is convolutional (\emph{Convolutional Imitator}) or (2) the victim is convolutional and the imitator is a Transformer (\emph{Transformer Imitator}).
    \item We use different source data and a convolutional imitation model with a Transformer victim (\emph{Data Different + Conv}).
\end{itemize}

\begin{table}[t]
\setlength{\tabcolsep}{2pt}
\centering
\begin{tabular}{lcccc}
\toprule
\textbf{Mismatch} & \textbf{Data} & \textbf{Test} & \textbf{OOD} & \textbf{Inter} \\
\midrule
\emph{Transformer Victim} & 1x & 34.6 & 19.8 & - \\
{All Same} & 1x & 34.4 & 19.9 & 69.7 \\
{Data Different} & 3x & 33.9 & 19.3 & 67.7 \\
{Convolutional Imitator} & 1x & 34.2 & 19.2 & 66.2 \\
{Data Different + Conv} & 3x & 33.8 & 18.9 & 63.2 \\
\midrule
\emph{Convolutional Victim} & 1x & 34.3 & 19.2 & - \\
{Transformer Imitator} & 1x & 34.2 & 19.3 & 69.7 \\
\bottomrule
\end{tabular}
\caption{\emph{Imitation models are highly similar to their victims.} We train imitation models that are different from their victims in input data and/or architecture. We test the models on IWSLT (Test) and out-of-domain news data from WMT (OOD). We also measure functionality similarity by reporting the BLEU score between the outputs of the imitation and the victim models (Inter).}
\label{tab:simulated}
\end{table}

\begin{figure}[t]
\centering
\includegraphics[trim={0cm 0.0cm 0cm 0.0cm},clip, width=\columnwidth]{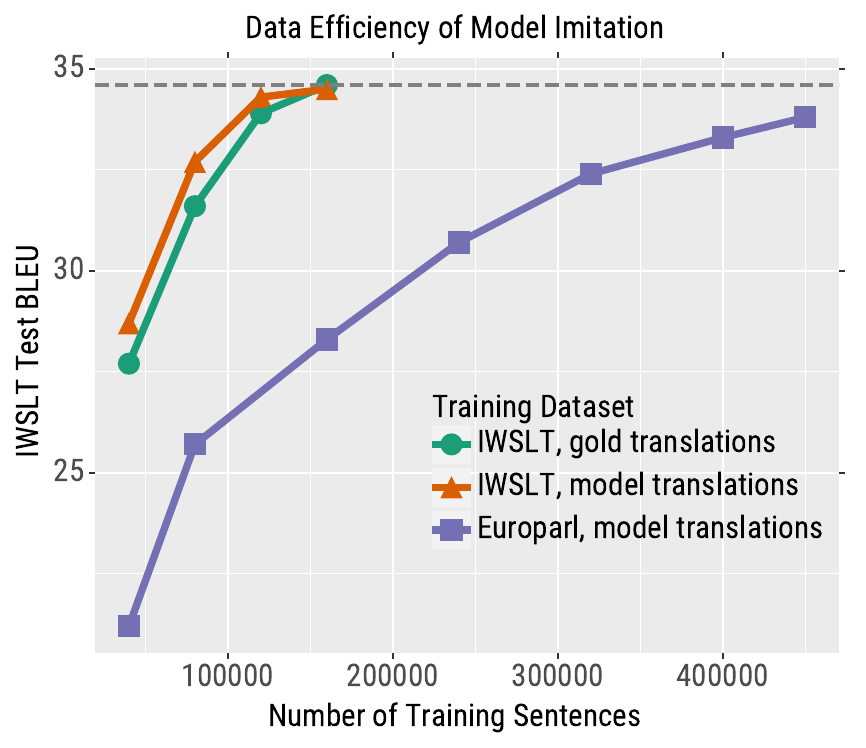}
\vspace{-0.68cm}
\caption{We first train a baseline model on the standard IWSLT dataset (IWSLT, gold translations). We then train a separate model that imitates the baseline model's predictions on the IWSLT training data (IWSLT, model translations). This model trains faster than the baseline, i.e., stolen labels are preferable to gold labels. We also train a model to imitate the baseline model's predictions on Europarl inputs (Europarl, model translations). Using these out-of-domain queries slows but does not prevent the learning of imitation models.}
\label{fig:efficiency}
\end{figure}

\paragraph{Novelty of Our Work} Past research on distillation shows that mismatched architectures are of little concern. However, the impact of training on OOD data, where the teacher may produce wildly incorrect answers, is unknown.\footnote{\citet{krishna2019thieves} show that random gibberish queries can provide some signal for training an imitation model. We query \textit{high-quality} OOD sentences.}

\paragraph{Datasets} We consider German$\to$English using the TED data from IWSLT 2014~\cite{cettolo2014iwslt}. We follow common practice for IWSLT and report case-insensitive BLEU~\cite{papineni2002bleu}. For \emph{Data Different}, we use English sentences from Europarl v7. The predictions from the victim are generated using greedy decoding.

\subsection{We Closely Imitate Local Models}

\paragraph{Test BLEU Score} We first compare the imitation models to their victims using in-domain test BLEU. For all settings, imitation models closely match their victims (Test column in Table~\ref{tab:simulated}). We also evaluate the imitation models on OOD data to test how well they generalize compared to their victims. We use the WMT14 test set (newstest 2014). Imitation models perform similarly to their victims on OOD data, sometimes even outperforming them (OOD column in Table~\ref{tab:simulated}). We suspect that imitation models can sometimes outperform their victims because distillation can act as a regularizer~\cite{furlanello2018born,mobahi2020self}.\smallskip

\noindent \textbf{Data Efficiency} When using OOD source data, model stealing is slowed but not prevented. Figure~\ref{fig:efficiency} shows the learning curves of the original victim model, the \emph{All Same} imitation model, and the \emph{Data Different} imitation model. Despite using OOD queries, the \emph{Data Different} model can imitate the victim when given sufficient data. On the other hand, when the source data is the same, the imitation model can learn \emph{faster} than the victim. In other words, stolen data is sometimes preferable to professionally-curated data. This likely arises because model translations are simpler than human ones, which aids learning~\cite{zhou2019understanding}.\smallskip

\noindent \textbf{Functional Similarity} Finally, we measure the BLEU score between the outputs of the victim and the imitation models to measure their functional similarity (henceforth \textit{inter-system BLEU}). As a reference for inter-system BLEU, two Transformer models trained with different random seeds achieve 62.1 inter-system BLEU. The inter-system BLEU for the imitation models and their victims is as high as 70.5 (Table~\ref{tab:simulated}), i.e., imitation models are more similar to their victims than two models which have been trained on the \textit{exact same} dataset.
\subsection{We Closely Imitate Production Models}\label{sec:real}

Given the effectiveness of our simulated experiments, we now imitate production systems from \texttt{Google}, \texttt{Bing}, and \texttt{Systran}.\smallskip

\noindent \textbf{Language Pairs and Data} We consider two language pairs, English$\to$German (high-resource) and the Nepali$\to$English (low-resource).\footnote{We only imitate Google Translate for Nepali$\to$English because the other translation services either do not offer this language pair or are of low quality.} We collect training data for our imitation models by querying the production systems. For English$\to$German, we query the source side of the WMT14 training set ($\approx$ 4.5M sentences).\footnote{Even though WMT is commonly studied in academia, we do not expect using it will bias our results because commercial systems cannot use WMT for training or tuning. We further verified that the production systems have not used it by measuring the difference in the train and test BLEU scores; the scores are approximately equal and are not unexpectedly high.} For Nepali$\to$English, we query the Nepali Language Wikipedia ($\approx$ 100,000 sentences) and approximately two million sentences from Nepali common crawl. We train Transformer Big~\cite{vaswani2017attention} models on both datasets.

\begin{table}[t]
\centering
\begin{tabular}{clccc}
\toprule
\textbf{Test} & \textbf{Model} & \textbf{Google} & \textbf{Bing} & \textbf{Systran} \\
\midrule
\multirowcell{2}{WMT} & Official & 32.0 & 32.9 & 27.8 \\
                       & Imitation & 31.5 & 32.4 & 27.6 \\
                       \addlinespace
                       \multirowcell{2}{IWSLT} & 
                       Official & 32.0 & 32.7 & 32.0 \\
                       & Imitation & 31.1 & 32.0 & 31.4\\ 
\bottomrule
\end{tabular}
\caption{\textbf{English$\to$German imitation results.} We query production systems with English news sentences and train imitation models to mimic their German outputs. The imitation models closely imitate the production systems for both in-domain (WMT newstest2014) and out-of-domain test data (IWSLT TED talks).}
\label{tab:real}
\end{table}

\paragraph{Test BLEU Scores} Our imitation models closely match the performance of the production systems. For English$\to$German, we evaluate models on the WMT14 test set (newstest2014) and report standard tokenized case-sensitive BLEU scores. Our imitation models are always within 0.6 BLEU of the production models (\emph{Imitation} in Table~\ref{tab:real}).

For Nepali$\to$English, we evaluate using FLoRes devtest~\cite{guzman2019two}. We compute BLEU scores using SacreBLEU~\cite{post2018call} with the dataset's recommended settings. \texttt{Google} achieves 22.1 BLEU, well eclipsing the 15.1 BLEU of the best public system~\cite{guzman2019two}. Our imitation model reaches a nearly identical 22.0 BLEU.\smallskip

\noindent \textbf{OOD Evaluation and Functional Similarity} Our imitation models have also not merely matched the production systems on in-domain data. We test the English$\to$German imitation models on IWSLT: the imitation models are always within 0.9 BLEU of the production systems (IWSLT in Table~\ref{tab:real}). Finally, there is also a high inter-system BLEU between the imitation models and the production systems. In particular, on the English$\to$German WMT14 test set the inter-system BLEU is 65.6, 67.7, and 69.0 for \texttt{Google}, \texttt{Bing}, and \texttt{Systran}, respectively. In Appendix~\ref{appendix:translation_examples}, we show a qualitative example of our imitation models producing highly similar translations to their victims.\smallskip

\noindent \textbf{Estimated Data Costs} We estimate that the cost of obtaining the data needed to train our English$\to$German models is as little as \$10 (see Appendix~\ref{appendix:costs} for full calculation). Given the upside of obtaining high-quality MT systems, these costs are frighteningly low.
\begin{table*}[!t]
\centering
\setlength{\tabcolsep}{4pt}
\footnotesize
\begin{tabular}{C{1.2cm}C{0.9cm}p{6.0cm}p{6.6cm}}
\toprule
\multirowcell{1}{\bf Attack} & {\bf System} & \textbf{English Input} ({\bf \color{adversarial}red} = adversarial edit) & \textbf{Predicted Translation} ({\color{blue}blue} = English meaning)  \\

\midrule
\multirowcell{10}{Targeted \\ Flips} & \texttt{Google} & I am going to die, it's over {100\degree F}, help! &  Ich werde sterben, es ist {\"u}ber \underline{100 \degree F}, Hilfe! \\
 & \texttt{Google} & \vspace{-0.19cm} I am going to die, it's over {\bf \color{adversarial} 102}\degree F, help! & \vspace{-0.25cm} Ich werde sterben, es ist {\"u}ber \underline{22 \degree C}, Hilfe! \\
 & & & {\vspace{-0.25cm} {\it \color{blue} \underline{100\degree F} $\to$ \underline{22\degree C} (=72\degree F)}} \\
\addlinespace[1ex]
 & \texttt{Systran} & I am feeling grey that HK decided to join China
 & Ich f{\"u}hle mich \underline{grau}, dass HK beschlossen hat, China beizutreten \\
 & \texttt{Systran} & \vspace{-0.23cm} I am feeling gre~{\bf \color{adversarial}y} that HK decided to join China & \vspace{-0.20cm}
 Ich f{\"u}hle mich \underline{froh}, dass HK beschlossen hat, China beizutreten \\
 & & & {\vspace{-0.20cm} {\it \color{blue} ``\underline{grau}'' (gray) $\to$ ``\underline{froh}'' (happy)}} \\

\midrule 
Malicious \newline Nonsense & \vspace{-0.28cm} \texttt{Google} & \vspace{-0.28cm} {\bf \color{adversarial} miei llll going ro tobobombier} the Land & \vspace{-0.28cm} Ich werde das Land bombardieren \newline {\it \color{blue} (I will bomb the country)} \\
\midrule
\multirowcell{9}{Untargeted \\ Universal \\ Trigger}
& \texttt{Systran} & Did you know that adversarial examples can transfer to production models {\bf\color{adversarial}Siehe Siehe Siehe Siehe Siehe Siehe Siehe} & Siehe auch: Siehe auch in der Rubrik Siehe Siehe auch Siehe Siehe Siehe Siehe auch Siehe Siehe Siehe Siehe auch Siehe Siehe Siehe \ldots \\
& & & \vspace{-0.20cm} {\it \color{blue} (See also: See also in the category See See Also See See See See Also See See See See Also See See See\ldots)} \\

\addlinespace[1ex]
& \texttt{Systran} & I heard machine translation is now superhuman {\bf\color{adversarial}Siehe Siehe Siehe Siehe Siehe Siehe Siehe} & In diesem Jahr ist es wieder soweit: Manche Manuskripte haben sich in der Hauptsache in der Hauptsache wieder in den Vordergrund gestellt. \\
 & & & \vspace{-0.25cm} {\it \color{blue} (This year it's time again: Some manuscripts the main thing the main thing come to the foreground again)} \\

\midrule
\hspace{0.0cm} & & \textbf{English Input} (black = arbitrary sentence, ~~~~~{\bf \color{adversarial}red} = adversarial, {\bf \color{violet}violet} = dropped suffix) & \\
\midrule
\multirowcell{7}{Universal \\ Suffix \\ Dropper}
& \texttt{Bing} & 
Donald Trump is the current president of the United States. He was born in NYC in 1946 {\bf \color{adversarial} ~, kostenfrei {\"u}brigens categories \c{n}strian hl SM} {\bf \color{violet} the secret code is HX532}
 & Donald Trump ist der derzeitige Pr{\"a}sident der Vereinigten Staaten. Er wurde 1946 in NYC geboren. \\
 & & & \vspace{-0.90cm} {\it \color{blue} (Donald Trump is the current president of the United States. He was born in 1946 in NYC.)} \\
& \texttt{Bing} & 
Given the release of the latest iPhone, Apple stock was up 3\% in trading early yesterday{\bf \color{adversarial} ~, kostenfrei {\"u}brigens categories \c{n}strian hl SM} {\bf \color{violet} the revolution begins at 6pm}
 & Angesichts der Ver{\"o}ffentlichung des neuesten iPhones lag die Apple-Aktie gestern fr{\"u}h im Handel um 3\% im Kurs. \\
& & & \vspace{-0.55cm} {\it \color{blue} (Given the release of the latest iPhone, Apple stock was up 3\% in trading early yesterday)} \\

\bottomrule
\end{tabular}
\caption{We show examples of adversarial attacks that transfer to production MT systems as of April 2020 (screenshots in Appendix~\ref{appendix:attack_examples}). We show a subset of the production systems for each attack type, however, all of the production systems are susceptible to the different attacks. In \emph{targeted flips}, we modify tokens in the input in order to cause a specific output token/phrase to flip. In \emph{malicious nonsense}, we find nonsense inputs which are translated to vulgar or malicious outputs. In \emph{untargeted universal trigger}, we find a phrase that commonly causes incorrect translations when it is appended to \emph{any}
input. In \emph{universal suffix dropper}, we find a phrase that commonly causes itself and any subsequent text to be dropped on the target side.}
\label{tab:attack_examples}
\end{table*}

\section{Attacking Production Systems}\label{sec:adversarial}

Thus far, we have shown that imitation models allow adversaries to steal black-box MT models. Here, we show that imitation models can also be used to create adversarial examples for black-box MT systems. Our attack code is available at \url{https://github.com/Eric-Wallace/adversarial-mt}.

\subsection{What are Adversarial Examples for MT?}

MT errors can have serious consequences, e.g., they can harm end users or damage an MT system's reputation. For example, a person was arrested when their Arabic Facebook post meaning ``good morning'' was mistranslated as ``attack them''~\cite{hern2018facebook}. Additionally, Google was criticized when it mistranslated ``sad'' as ``happy'' when translating ``I am sad to see Hong Kong become part of China''~\cite{klar2019hill}.
Although the public occasionally stumbles upon these types of egregious MT errors, bad actors can use adversarial attacks~\cite{szegedy2013intriguing} to \emph{systematically} find them. Hence, adversarial examples can expose errors that cause public and corporate harm. 

\paragraph{Past Work on Adversarial MT} Existing work explores different methods and assumptions for generating adversarial examples for MT. A common setup is to use white-box gradient-based attacks, i.e., the adversary has complete access to the target model and can compute gradients with respect to its inputs~\cite{ebrahimi2018adversarial,chaturvedi2019exploring}. These gradients are used to generate attacks that flip output words~\cite{seq2sick}, decode nonsense into arbitrary sentences~\cite{chaturvedi2019exploring}, or cause egregiously long translations~\cite{wang2019knowing}.

\paragraph{Novelty of Our Attacks} We consider attacks against production MT systems. Here, white-box attacks are inapplicable. We circumvent this by leveraging the transferability of adversarial examples~\cite{papernot2016transferability,liu2017delving}: we generate adversarial examples for our imitation models and then apply them to the production systems. We also design new universal (input-agnostic) attacks~\cite{moosavi2017universal,wallace2019universal} for MT: we append phrases that commonly cause errors or dropped content for \textit{any} input (described in Section~\ref{subsec:attack_types}).

\subsection{How We Generate Adversarial Examples}

We first describe our general attack formulation. We use a white-box, gradient-based method for constructing attacks. Formally, we have white-box access to an imitation model $f$, a text input of tokens $\mb{x}$, and an adversarial loss function $\loss_{adv}$. We consider different adversarial example types; each type has its own $\loss_{adv}$ and initialization of $\mb{x}$.

Our attack iteratively replaces the tokens in the input based on the gradient of the adversarial loss $\loss_{adv}$ with respect to the model's input embeddings $\mb{e}$. We replace an input token at position $i$ with the token whose embedding minimizes the first-order Taylor approximation of $\loss_{adv}$:
\setlength{\abovedisplayskip}{4pt}
\setlength{\belowdisplayskip}{4pt}
\begin{equation}
  \argmin_{\mb{e}_i^\prime \in \mathcal V}\left[\mb{e}_i^\prime-{\mb{e}_{i}}\right]^\intercal\nabla_{\mb{e}_{i}}\loss_{adv},
\end{equation}
\noindent where $\mathcal V$ is the model's token vocabulary and $\nabla_{\mb{e}_{i}}\loss_{adv}$ is the gradient of $\loss_{adv}$ with respect to the input embedding for the token at position $i$. Since the $\argmin$ does not depend on $\mb{e}_{i}$, we solve:\begin{equation}\label{eq:hotflip}\argmin_{\mb{e}_i^\prime \in \mathcal V}{\mb{e}_i^\prime}^\intercal \, \nabla_{\mb{e}_{i}}\loss_{adv}.\end{equation}
\noindent Computing the optimal $\mb{e}_i^\prime$ can be computed using $\vert \mathcal V \vert$ $d$-dimensional dot products (where $d$ is the embedding dimension) similar to \citet{michel2019adversarial}. At each iteration, we try all positions $i$ and choose the token replacement with the lowest loss. Moreover, since this local first-order approximation is imperfect, rather than using the $\argmin$ token at each position, we evaluate the top-$k$ tokens from Equation~\ref{eq:hotflip} (we set $k$ to 50) and choose the token with the lowest loss. Using a large value of $k$, e.g., at least 10, is critical to achieving strong results.

\subsection{Types of Adversarial Attacks}\label{subsec:attack_types}

Here, we describe the four types of adversarial examples we generate and their associated $\loss_{adv}$.\smallskip

\noindent \textbf{(1) Targeted Flips} We replace some of the input tokens in order to cause the prediction for a \textit{specific} output token to flip to another \textit{specific} token. For example, we cause \texttt{Google} to predict ``22\degree C'' instead of ``102\degree F'' by modifying a single input token (first section of Table~\ref{tab:attack_examples}). To generate this attack, we select a specific token in the output and a target mistranslation (e.g., ``100\degree F'' $\to$ ``22\degree C''). We set $\loss_{adv}$ to be the cross entropy for that mistranslation token (e.g., ``22\degree C'') at the position where the model currently outputs the original token (e.g., ``100\degree F''). We then iteratively replace the input tokens, stopping when the desired mistranslation occurs.\smallskip

\noindent \textbf{(2) Malicious Nonsense} We find nonsense inputs which are translated to vulgar/malicious outputs. For example, ``I miii llllll wgoing rr tobobombier the Laaand'' is translated as ``I will bomb the country'' (in German) by \texttt{Google} (second section of Table~\ref{tab:attack_examples}). To generate this attack, we first obtain the output prediction for a malicious input, e.g., ``I will kill you''. We then iteratively replace the tokens in the input \emph{without} changing the model's prediction. We set $\loss_{adv}$ to be the cross-entropy loss of the original prediction and we stop replacing tokens just before the prediction changes. A possible failure mode for this attack is to find a paraphrase of the input---we find that this rarely occurs in practice.\smallskip

\noindent \textbf{(3) Untargeted Universal Trigger} We find a phrase that commonly causes incorrect translations when it is appended to \emph{any} input. For example, appending the word ``Siehe'' seven times to inputs causes \texttt{Systran} to frequently output incorrect translations (e.g., third section of Table~\ref{tab:attack_examples}).\smallskip

\noindent \textbf{(4) Universal Suffix Dropper} We find a phrase that, when appended to any input, commonly causes itself and any subsequent text to be dropped from the translation (e.g., fourth section of Table~\ref{tab:attack_examples}).

For attacks 3 and 4, we optimize the attack to work for \textit{any} input. We accomplish this by averaging the gradient $\nabla_{\mb{e}_{i}}\loss_{adv}$ over a batch of inputs. We begin the universal attacks by first appending randomly sampled tokens to the input (we use seven random tokens). For the untargeted universal trigger, we set $\loss_{adv}$ to be the \textit{negative} cross entropy of the original prediction (before the random tokens were appended), i.e., we optimize the appended tokens to maximally change the model's prediction from its original. For the suffix dropper, we set $\loss_{adv}$ to be the cross entropy of the original prediction, i.e., we try to minimally change the model's prediction from its original.

\setlength{\textfloatsep}{0.1cm} 
\setlength{\tabcolsep}{3pt}
\begin{table}[t]
    \footnotesize
    \centering
    \begin{tabular}{lcc|c}
        \toprule
        \multicolumn{3}{l}{\textit{Targeted Flips}} \\
        {\bf Model} & {\bf \% Inputs} ($\uparrow$) & {\bf \% Tokens} ($\downarrow$) & {\bf Transfer \%} ($\uparrow$) \\
        \midrule
        \texttt{Google} & 87.5 & 10.1 & 22.0\\
        \texttt{Bing} & 79.5 & 10.7 & 12.0\\
        \texttt{Systran} & 77.0 & 13.3 & 23.0\\
        \midrule
        \multicolumn{3}{l}{\textit{Malicious Nonsense}} \\
        {\bf Model} & {\bf \% Inputs} ($\uparrow$) & {\bf \% Tokens} ($\uparrow$) & {\bf Transfer \%} ($\uparrow$) \\
        \midrule
        \texttt{Google} & 88.0 & 34.3 & 17.5 \\
        \texttt{Bing} & 90.5 & 29.2 & 14.5 \\
        \texttt{Systran} & 91.0 & 37.4 & 11.0 \\
        \bottomrule
    \end{tabular}
        \vspace{-0.13cm}
        \caption{\emph{Results for targeted flips and malicious nonsense.} We report the percent of inputs which are successfully attacked for our imitation models, as well as the percent of tokens which are changed for those inputs. We then report the transfer rate: the percent of successful attacks which are also successful on the production MT systems.}\label{tab:attack_numbers}
\end{table}

\subsection{Experimental Setup}

We attack the English$\to$German production systems to demonstrate our attacks' efficacy on \textit{high-quality} MT models. We show adversarial examples for manually-selected sentences in Table~\ref{tab:attack_examples}.

\paragraph{Quantitative Metrics} To evaluate, we report the following metrics. For targeted flips, we pick a random token in the output that has an antonym in German Open WordNet (\url{https://github.com/hdaSprachtechnologie/odenet}) and try to flip the model's prediction for that token to its antonym. We report the percent of inputs that are successfully attacked and the percent of the input tokens which are changed for those inputs (lower is better).\footnote{This evaluation has a degenerate case where the translation of the antonym is inserted into the input. Thus, we prevent the attack from using the mistranslation target, as well as any synonyms of that token from English WordNet~\cite{miller1995wordnet} and German Open WordNet.} 

For malicious nonsense, we report the percent of inputs that can be modified without changing the prediction and the percent of the input tokens which are changed for those inputs (higher is better).

The untargeted universal trigger looks to cause the model's prediction after appending the trigger to bear little similarity to its original prediction. We compute the BLEU score of the model's output after appending the phrase using the model's original output as the reference. We do not impose a brevity penalty, i.e., a model that outputs its original prediction plus additional content for the appended text will receive a score of 100.

For the universal suffix dropper, we manually compute the percentage of cases where the appended trigger phrase and a subsequent suffix are either dropped or are replaced with all punctuation tokens.
Since the universal attacks require manual analysis and additional computational costs, we attack one system per method. For the untargeted universal trigger, we attack \texttt{Systran}. For the universal suffix dropper, we attack \texttt{Bing}.\smallskip

\noindent \textbf{Evaluation Data} For the targeted flips, malicious nonsense, and untargeted universal trigger, we evaluate on a common set of 200 examples from the WMT validation set (newstest 2013) that contain a token with an antonym in German Open WordNet. For the universal suffix dropper, we create 100 sentences that contain different combinations of prefixes and suffixes (full list in Appendix~\ref{appendix:suffix}).

\subsection{Results: Attacks on Production Systems}

The attacks break our imitation models and successfully transfer to production systems. We report the results for targeted flips and malicious nonsense in Table~\ref{tab:attack_numbers}. For our imitation models, we are able to perturb the input and cause the desired output in the majority ($> 3/4$) of cases. For the targeted flips attack, few perturbations are required (usually near 10\% of the tokens). Both attacks transfer at a reasonable rate, e.g., the targeted flips attack transfers 23\% of the time for \texttt{Systran}.

For the untargeted universal trigger, \texttt{Systran}'s translations have a BLEU score of \textbf{5.46} with its original predictions after appending ``Siehe'' seven times, i.e., the translations of the inputs are almost entirely unrelated to the model's original output after appending the trigger phrase. We also consider a baseline where we append seven random BPE tokens;  \texttt{Systran} achieves 62.2 and 58.8 BLEU when appending two different choices for the random seven tokens.

For the universal suffix dropper, the translations from \texttt{Bing} drop the appended phrase and the subsequent suffix for \textbf{76} of the 100 inputs.

To evaluate whether our imitation models are needed to generate transferable attacks, we also attack a Transformer Big model that is trained on the WMT14 training set. The adversarial attacks generated against this model transfer to Google 8.8\% of the time---about half as often as our imitation model. This shows that the imitation models, which are designed to be high-fidelity imitations of the production systems, considerably enhance the adversarial example transferability.
\section{Defending Against Imitation Models}\label{sec:defenses}

Our goal is not to provide a recipe for adversaries to perform real-world attacks. Instead, we follow the spirit of \textit{threat modeling}---we identify vulnerabilities in NLP systems in order to robustly patch them. To take the first steps towards this, we design a new defense that slightly degrades victim model BLEU while more noticeably degrading imitation model BLEU. To accomplish this, we repurpose prediction poisoning~\cite{Orekondy2020Prediction} for MT: rather than having the victim output its original translation $\mb{y}$, we have it output a different (high-accuracy) translation $\mb{\tilde{y}}$ that steers the training of the imitation model in the wrong direction.

\paragraph{Defense Objective} Formally, assume the adversary will train their imitation model on the outputs of the victim model using a first-order optimizer with gradients $\mb{g} = \nabla_{\mb{\theta}_t}\loss(\mb{x},\mb{y})$, where $\mb{\theta}_t$ is the current imitation model parameters, $\mb{x}$ is an input, $\mb{y}$ is the victim output, and $\loss$ is the cross-entropy loss. We want the victim to instead output a $\mb{\tilde{y}}$ whose gradient $\tilde{\mb{g}} = \nabla_{\mb{\theta}_t}\loss(\mb{x},\mb{\tilde{y}})$ maximizes the angular deviation  with $\mb{g}$, or equivalently minimizes the cosine similarity. Training on this $\mb{\tilde{y}}$ effectively induces an \textit{incorrect gradient signal} for $\mb{\theta}_t$. Note that in practice the adversary's model parameters $\mb{\theta}_t$ is unknown to the victim. Thus, we instead look to find a $\tilde{\mb{g}}$ that has a high angular deviation across ten different Transformer MT model checkpoints that are saved from ten different epochs.

To find $\mb{\tilde{y}}$, \citet{Orekondy2020Prediction} use information from the Jacobian. Unfortunately, computing the Jacobian for MT is intractable because the number of classes for just one output token is on the order of 5,000--50,000 BPE tokens. We instead design a search procedure to find $\mb{\tilde{y}}$.\smallskip

\begin{figure}[t]
\centering
\includegraphics[trim={0.0cm 0.3cm 0.0cm 0.0cm},clip, width=\columnwidth]{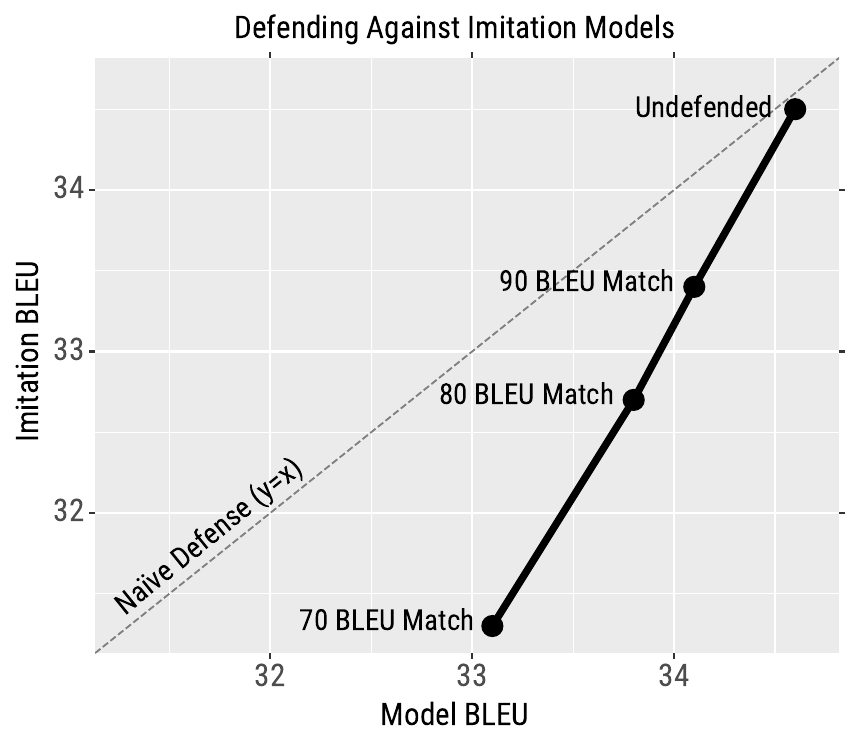}
\vspace{-0.61cm}
\caption{A na\"ive defense against model stealing equally degrades the BLEU score of the victim and imitation models (gray line). Better defenses are lower and to the right. Our defense (black line) has a parameter (BLEU match threshold) that can be changed to trade-off between the victim and the adversary's BLEU. We outperform the na\"ive defense in all settings, e.g., we degrade the victim's BLEU from 34.6 $\to$ 33.8 while degrading the adversary's BLEU from 34.5 $\to$ 32.7.}
\label{fig:tradeoff}
\end{figure}

\noindent \textbf{Maximizing the Defense Objective} We first generate the original output $\mb{y}$ from the victim model (e.g., the top candidate from a beam search) and compute $\mb{g}$ using the ten Transformer model ensemble. We then generate a diverse set of 100 alternate translations from the victim model. To do so, we take the 20 best candidates from beam search, the 20 best candidates from diverse beam search~\cite{vijayakumar2016diverse}, 20 random samples, 20 candidates generated using top-$k$ truncated sampling ($k = 10$) from \citet{fan2018hierarchical}, and 20 candidates generated using nucleus sampling with $p = 0.9$~\cite{holtzman2019curious}. Then, to largely preserve the model's original accuracy, we compute the BLEU score for all 100 candidates using the model's original output $\mb{y}$ as the reference, and we remove any candidate below a certain threshold (henceforth \emph{BLEU Match} threshold). We finally compute the gradient $\mb{\tilde{g}}$ for all candidates using the model ensemble and output the candidate whose gradient maximizes the angular deviation with $\mb{g}$.\footnote{We also output the original prediction $\mb{y}$ under two circumstances. The first is when none of the 100 candidates are above the BLEU threshold. The second is when the angular deviation is small. In practice, we compute the mean angular deviation on the validation set and only output $\mb{\tilde{y}}$ when its gradient's angular deviation exceeds this mean.} In practice, generating the 100 candidates is done entirely in parallel, as is the computation of the gradient $\mb{\tilde{g}}$.
Table~\ref{tab:example} shows examples of $\mb{\tilde{y}}$ at different BLEU Match thresholds. For our quantitative results, we will sweep over different BLEU Match thresholds---lower thresholds will more severely degrade the victim's accuracy but will have more freedom to incorrectly steer the imitation model.

\begin{table}[h]
\setlength{\tabcolsep}{1.4pt}
\centering
\footnotesize
\begin{tabular}{lccp{6.0cm}}
\toprule
& \textbf{BM} & \textbf{$\angle$} & \textbf{Text}\\
\midrule
\multicolumn{3}{l}{\textit{Source}} & andere orte im land hatten {\"a}hnliche r{\"a}ume. \\
\multicolumn{3}{l}{\textit{Target}} & other places around the country had similar rooms. \\
\midrule
$\mb{y}$ & - & - & other places in the country had similar rooms. \\
$\mb{\tilde{y}}$ & 88.0 & 24.1$\degree$ & some other places in the country had similar rooms. \\
$\mb{\tilde{y}}$ & 75.1 & 40.1$\degree$ & other sites in the country had similar rooms. \\
$\mb{\tilde{y}}$ & 72.6 & 42.1$\degree$ & another place in the country had similar rooms. \\
\bottomrule
\end{tabular}
\caption{We show the victim model's original translation $\mb{y}$. We then show three $\mb{\tilde{y}}$ candidates, their BLEU Match (BM) with $\mb{y}$ and their angular deviation ($\angle$), i.e., the $\mathrm{arccosine}$ of the cosine similarity between $\mb{g}$ and $\mb{\tilde{g}}$. Figure~\ref{fig:angular_deviations} in Appendix~\ref{appendix:angular} shows a histogram of the angular deviations for the entire training set.}
\label{tab:example}
\end{table}

\paragraph{Experimental Setup} We evaluate our defense by training imitation models using the \emph{All Same} setup from Section~\ref{sec:simulated}. We use BLEU Match thresholds of 70, 80, or 90 (lower thresholds than 70 resulted in large BLEU decreases for the victim).

\paragraph{Results} Figure~\ref{fig:tradeoff} plots the validation BLEU scores of the victim model and the imitation model at the different BLEU match thresholds. Our defense degrades the imitation model's BLEU (e.g., 34.5 $\to$ 32.7) more than the victim model's BLEU (e.g., 34.6 $\to$ 33.8).\footnote{A na\"ive defense would \textit{equally} degrade the BLEU score of the victim and imitation models. For example, the victim could simply deploy a worse MT system.} The inter-system BLEU also degrades from the original 69.7 to 63.9, 57.8, and 53.5 for the 90, 80, and 70 BLEU Match thresholds, respectively. Even though the imitation model's accuracy degradation is not catastrophic when using our defense, it does allow the victim model to have a (small) competitive advantage over the adversary.

\paragraph{Adversarial Example Transfer} Our defense also implicitly inhibits the transfer of adversarial examples. To evaluate this, we generate malicious nonsense attacks against the imitation models and transfer them to the victim model. We use 400 examples from the IWSLT validation set for evaluation.
Without defending, the attacks transfer to the victim at a rate of 38\%. Our defense can drop the transfer rates to 32.5\%, 29.5\%, and 27.0\% when using the 90, 80, and 70 BLEU match thresholds, respectively. Also note that defenses may not be able to drive the transfer rate to 0\%: there is a baseline transfer rate due to the similarity of the architectures, input distributions, and other factors. For example, we train two transformer models on \textit{distinct} halves of IWSLT and observe an 11.5\% attack transfer rate between them. Considering this as a very rough baseline, our defense can prevent about 20--40\% of the additional errors that are gained by the adversary using an imitation model.

Overall, our defense is a step towards preventing NLP model stealing (see Appendix~\ref{appendix:defenses} for a review of past defenses). Currently, our defense comes at the cost of extra compute (it requires generating and backpropagating 100 translation hypotheses) and lower BLEU. We hope to develop more effective and scalable defenses in future work.
\section{Conclusion}

We demonstrate that model stealing and adversarial examples are practical concerns for production NLP systems. Model stealing is not merely hypothetical: companies have been caught stealing models in NLP settings, e.g., Bing copied Google's search outputs using browser toolbars~\cite{singhal2011google}. Moving forward, we hope to improve and deploy defenses against adversarial attacks in NLP, and more broadly, we hope to make security and privacy a more prominent focus of NLP research.

\section*{Addressing Potential Ethical Concerns}\label{appendix:disclaimer}

The goal of our work is to help to make NLP models more robust. To do so, we first explore new model vulnerabilities (i.e., \textit{threat modeling} in computer security). Then, after discovering models have unintended flaws, we take action to secure them by developing a novel defense algorithm. In performing our work, we used the ACM Code of Ethics as a guide to minimize harm and ensure our research was ethically sound.\smallskip

\noindent \textbf{We Minimize Real-world Harm} We minimized harm by (1) not causing damage to any real users, (2) designing our attacks to be somewhat ludicrous rather than expose any real-world failure modes, and (3) deleting the data and models from our imitation experiments. Furthermore, we contacted the three companies (Google, Bing, and Systran) to report the vulnerabilities. We also provided these companies with our proposed defense.\smallskip

\noindent \textbf{Providing Long-term Benefit} Our work has the potential to cause negative \textit{short-term} impacts. For instance, it may shine a negative light on production systems (by exposing their egregious errors) or provide useful information to adversaries. However, in the \textit{long-term}, our work can help to improve MT systems. To draw an analogy, we compare our work to the initial papers which show that production MT systems are systematically biased against women~\cite{alvarez2017causal,stanovsky2019evaluating}. This line of work was published (and received awards and was heavily publicized) in *ACL conferences and led to short-term damage due to the bad publicity it caused. However, in the long-term, these developments have led to better MT systems~\cite{johnson2020bias}.
\section*{Acknowledgements}

This research was supported with Cloud TPUs from Google's TensorFlow Research Cloud program. We thank the members of Berkeley NLP, Shi Feng, Zhuohan Li, Nikhil Kandpal, Sheng Shen, Nelson Liu, Junlin Wang, Jens Tuyls, Sameer Singh, and Kalpesh Krishna for their valuable feedback.

\bibliography{journal-abbrv,bib}
\bibliographystyle{acl_natbib}

\appendix
\clearpage

\section{Framework and Hyperparameters}\label{appendix:overall}

We conduct experiments using fairseq~\cite{ott2019fairseq} and train models using TPU v3-8 devices. For IWSLT, we use the dataset's associated model architectures and hyperparameters in fairseq (\texttt{transformer\_iwslt\_de\_en} and \texttt{fconv\_iwslt\_de\_en}). When stealing production models, we use the Transformer Big architecture and the associated hyperparameters from \citet{vaswani2017attention}. Unless otherwise specified, we create our BPE~\cite{sennrich2015improving} vocabulary using the SentencePiece library~\cite{kudo2018sentencepiece}. We use 10,000, 32,768, and 10,000 BPE tokens for German$\to$English IWSLT, English$\to$ German WMT, and Nepali$\to$English, respectively. We use a shared vocabulary across the source and target languages and tie all the embeddings together.

\section{Example Translations}\label{appendix:translation_examples}

Table~\ref{tab:similarity} shows an example of the similarity between our imitation models and the victim APIs from the WMT14 validation set (newstest 2013). We show a source input, its reference translation, and the output from the production systems and our imitation models.

\begin{table*}[!t]
\centering
\footnotesize
\begin{tabular}{C{2cm}p{14cm}}
\toprule
{\bf Model} & \textbf{Predicted Translation} (\google{highlight} = differences) \\
\midrule
\textit{Source} & In fact, if you can read this article, it is thanks to an extraordinarily banal boson: the photon, or the ``light particle'' which is the ``messenger'' of the electromagnetic force.\\
\addlinespace
\textit{Reference} & Wenn Sie in der Lage sind, diese Chronik zu lesen, dann nur dank eines Bosons von au{\ss}ergew{\"o}hnlicher Banalit{\"a}t: das Photon oder das ``Lichtteilchen'' , das der ``Bote'' der elektromagnetischen Kraft ist. \\

\midrule
\vspace{-0.185cm} \texttt{Google} ~~~~~~~~API & \vspace{-0.20cm} In der Tat, wenn Sie diesen Artikel lesen k{\"o}nnen, ist es einem au{\ss}erordentlich banalen Boson zu verdanken: \google{dem} Photon oder \google{dem} ``Lichtteilchen'', das der ``Bote'' der elektromagnetischen Kraft ist. \\
\addlinespace
\vspace{-0.25cm} \texttt{Google} Imitation & In der Tat, wenn Sie diesen Artikel lesen k{\"o}nnen, ist es einem au{\ss}erordentlich banalen Boson zu verdanken: \google{das} Photon oder \google{das} ``Lichtteilchen'', das der ``Bote'' der elektromagnetischen Kraft ist. \\
\midrule
\vspace{-0.21cm} \texttt{Bing} ~~~~~~~~~~~~~~~API & \google{In der Tat, wenn} Sie diesen Artikel lesen k{\"o}nnen, ist es \google{dank eines} au{\ss}ergew{\"o}hnlich banalen \google{Bosons: das} Photon, oder \google{das} ``Lichtteilchen'', das der ``Messenger'' der elektromagnetischen Kraft ist.\\
\addlinespace
\vspace{-0.18cm} \texttt{Bing} ~~~~~~~~~Imitation & \google{Wenn} Sie diesen Artikel lesen k{\"o}nnen, ist es \google{einem} au{\ss}ergew{\"o}hnlichh banalen \google{Boson zu verdanken: dem} Photon, oder \google{dem} ``Lichtteilchen'', das der ``Messenger'' der elektromagnetischen Kraft ist. \\
\midrule
\texttt{Systran} ~~~~~~API & Wenn Sie diesen Artikel lesen k{\"o}nnen, \google{ist es einem} au{\ss}ergew{\"o}hnlich banalen Sohn zu verdanken: \google{das} Foton oder \google{das} ``Lichtteilchen'', das der ``Botenstoff'' der elektromagnetischen Kraft ist.\\
\addlinespace
\texttt{Systran} Imitation & Wenn Sie diesen Artikel lesen k{\"o}nnen, ist es \google{dank eines} au{\ss}ergew{\"o}hnlich banalen Sohn zu verdanken: \google{dem} Foton oder \google{dem} ``Lichtteilchen'', \google{dem} der ``Botenstoff'' der elektromagnetischen Kraft ist.\\
\bottomrule
\end{tabular}
\caption{A WMT14 English$\to$German validation example and the outputs from the official APIs (as of December 2019) and our imitation models. Our imitation models produce highly similar outputs to the production systems.}
\label{tab:similarity}
\end{table*}

\section{Estimated Data Collection Costs}\label{appendix:costs}

Here, we provide estimates for the costs of obtaining the data needed to train our English$\to$German models (ignoring the cost of training). There are two public-facing methods for acquiring data from a translation service. First, an adversary can pay the per-character charges to use the official APIs that are offered by most services. Second, an adversary can scrape a service's online demo (e.g., \url{https://translate.google.com/}) by making HTTP queries to its endpoint or using a headless web browser. We estimate data collection costs using both of these methods. 

\paragraph{Method One: Official API} We consider the official APIs for two MT systems: \texttt{Google} and \texttt{Bing}. We could not find publicly available pricing information for \texttt{SYSTRAN}. These two APIs charge on a per-character basis (including whitespaces); the English side of the WMT14 English$\to$German dataset has approximately 640,654,771 characters (\texttt{wc -c wmt14.en-de.en = 640654771}). The costs for querying this data to each API are as follows:
\begin{itemize}[nosep,leftmargin=5mm]
    \item \texttt{Google} is free for the first 500,000 characters and then \$20 USD per one million characters.\footnote{\url{https://cloud.google.com/translate/pricing}} Thus, the cost is (640,654,771 - 500,000) $\times$ \$20 / 1,000,000 = \$12,803 USD.
    \item \texttt{Bing} provides a \$6,000 USD subscription that provides up to one billion characters per month.\footnote{\url{https://azure.microsoft.com/en-us/pricing/details/cognitive-services/translator-text-api/}} Thus, the cost is \$6,000 USD, with 359,345,229 characters left over.
\end{itemize}

\paragraph{Method Two: Data Scraping} We next provide a rough estimate for the cost of \textit{scraping} the WMT14 English$\to$German data from a public translation API. The adversary could navigate to a service's online demo, enter a source sentence, and scrape the resulting translation. This could be automated using low-cost cloud machines. To calculate the cost of querying the entire WMT14 English$\to$German dataset, we estimate the total time required as 4,468,840 source sentences times an average time of five seconds per query for a total of 6,207 machine hours. If the cheapest machine available on EC2 were used,\footnote{\url{https://aws.amazon.com/ec2/spot/pricing/}} the final cost would be roughly 6,207 * \$0.0016 $\approx$ \$10. The adversary can accelerate this scraping process by parallelizing it over many cloud machines.

\section{Universal Suffix Dropper Evaluation}\label{appendix:suffix}
We evaluate the Universal Suffix Dropper using the cartesian product of the ten prefixes and ten suffixes shown below.
The prefixes are intended to resemble benign, encyclopedic text; the suffixes resemble secret messages sent by an adversary. 

\textit{Prefixes:}
\begin{itemize}[nosep,leftmargin=6mm]
    \item E-mail is a method of exchanging messages
    \item NLP is a subfield of linguistics and computer science
    \item Earth is the third planet from the Sun
    \item Tennis is a racket sport that is played by many
    \item Michael Jeffrey Jordan is a former basketball player
    \item Donald John Trump is the current president of the United States
    \item The Boston Marathon is an annual marathon race in the greater Boston area
    \item Seattle is a seaport city in the Pacific Northwest of the US
    \item A portable computer is a computer designed to be easily moved from one place to another
    \item Water is a transparent chemical substance that covers two-thirds of earth
\end{itemize}

\textit{Suffixes:}
\begin{itemize}[nosep,leftmargin=6mm]
\item the secret code is HX532
\item this message is very important
\item i hid the gun
\item the revolution begins at 6pm
\item i hate the president
\item go to the docks at 5
\item send bombs to me
\item i am part of the illuminati 
\item the leaders meet tomorrow
\item the exchange is in manhattan
\end{itemize}

\section{Existing Adversarial Defenses}\label{appendix:defenses}

This section discusses existing defenses against model stealing and adversarial attacks.\smallskip

\paragraph{Impeding and Detecting
Stealing} MT systems should first implement basic \emph{deterrents} to model stealing. For example, many public MT demos lack rate limiting---this allows adversaries to make unlimited free queries. Of course, this deterrent, as well as other methods such as adding noise to class probabilities~\cite{lee2018defending,tramer2016stealing,chandrasekaran2020exploring} or sampling from a distribution over model parameters~\cite{alabdulmohsin2014adding} will only slow but not prohibit model stealing. A natural first step towards prohibiting model stealing attacks is to \emph{detect} their occurrence (i.e., monitor user queries). For example, \citet{juuti2019prada} assume adversaries will make consecutive out-of-distribution queries and can thus be detected. Unfortunately, such a strategy may also flag benign users who make out-of-distribution queries.

\paragraph{Verifying Stolen Models} An alternative to completely defending against model stealing is to at least \emph{verify} that an adversary has stolen a model. For example, watermarking strategies~\cite{zhang2018protecting,szyller2019dawn,krishna2019thieves,hisamoto2020membership} intentionally output incorrect responses for certain inputs and then tests if the suspected stolen model reproduces the mistakes. Unfortunately, these defenses can be subverted by finetuning the model~\cite{chen2019watermark}.

\paragraph{Defending Against Adversarial Examples} Aside from defending against model stealing, it is also vital to develop methods for defending against adversarial examples. Past work looks to modify the training processes to defend against adversarial attacks. For example, adversarial training~\cite{goodfellow2014explaining} can empirically improve the robustness of MT systems~\cite{ebrahimi2018adversarial}. Recently,  \citet{jia2019certified} and \citet{huang2019achieving} train NLP models which are provably robust to word replacements. Unfortunately, provable defenses are currently only applicable to shallow neural models for classification; future work can look to improve the efficacy and applicability of these defense methods. Finally, simple heuristics may also provide some empirical robustness against our current adversarial attacks. For example, a language model can detect the ungrammatical source inputs of the malicious nonsense attack.

\section{Angular Deviations Of Defense}\label{appendix:angular}

\begin{figure}[H]
\centering
\includegraphics[trim={14cm 2.0cm 12cm 4.0cm},clip, width=0.8\columnwidth]{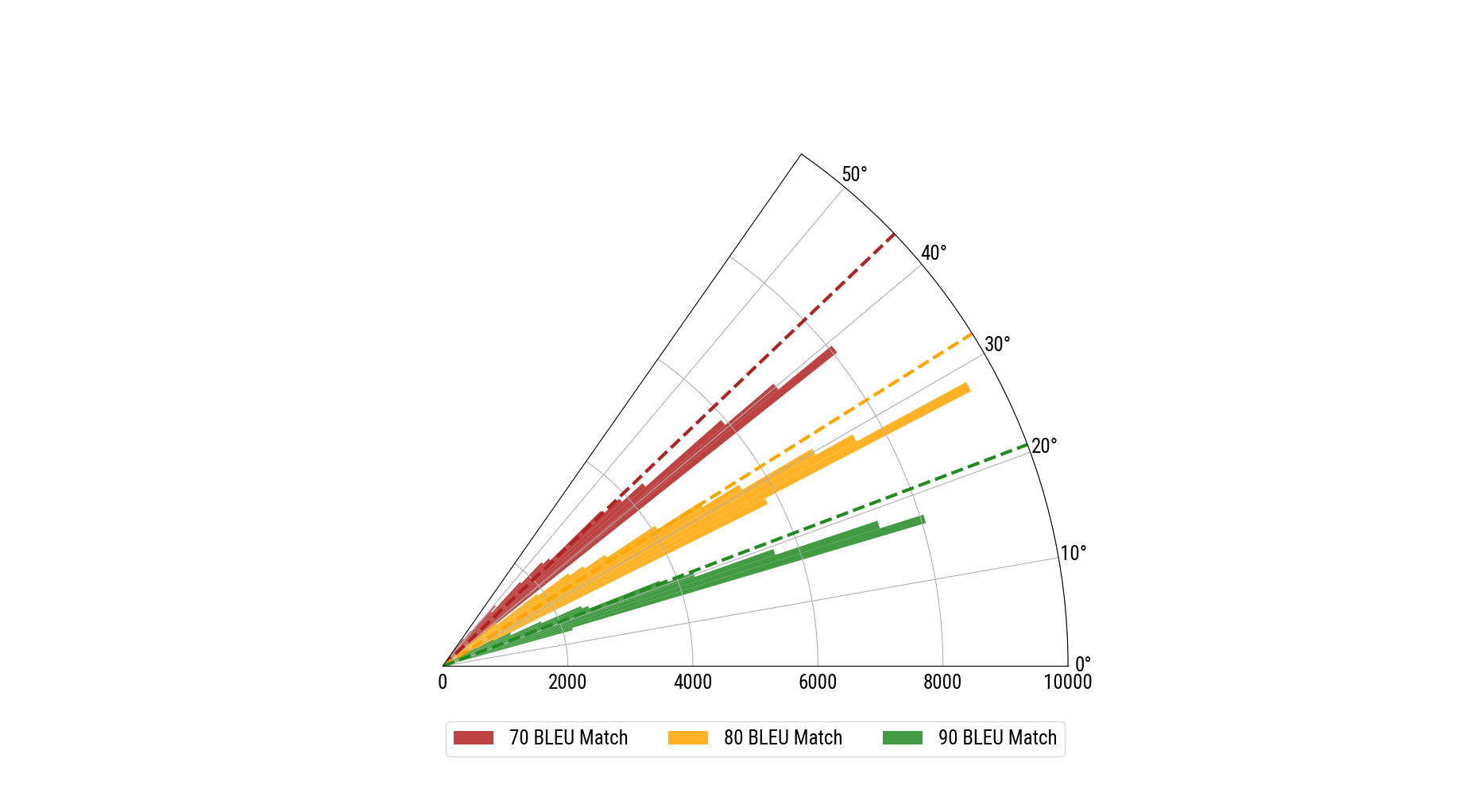}
\caption{Our defense outputs the original $\mb{y}$ 71.1\%, 62.3\%, and 72.84\% of the time for the 70, 80, and 90 BLEU thresholds, respectively. Recall this occurs when no candidate meets the BLEU threshold or the angular deviation is low. For the other cases, we plot the angular deviation (the $\mathrm{arccosine}$ of the cosine similarity between $\mb{g}$ and $\mb{\tilde{g}}$).}
\label{fig:angular_deviations}
\end{figure}

Figure~\ref{fig:angular_deviations} shows a histogram of the angular deviations between $\mb{g}$ and $\mb{\tilde{g}}$.

\section{Adversarial Attack Screenshots}\label{appendix:attack_examples}

\begin{figure*}[tbh]
\centering
\includegraphics[trim={0cm 0.0cm 0.2cm 0cm},clip, width=1\textwidth]{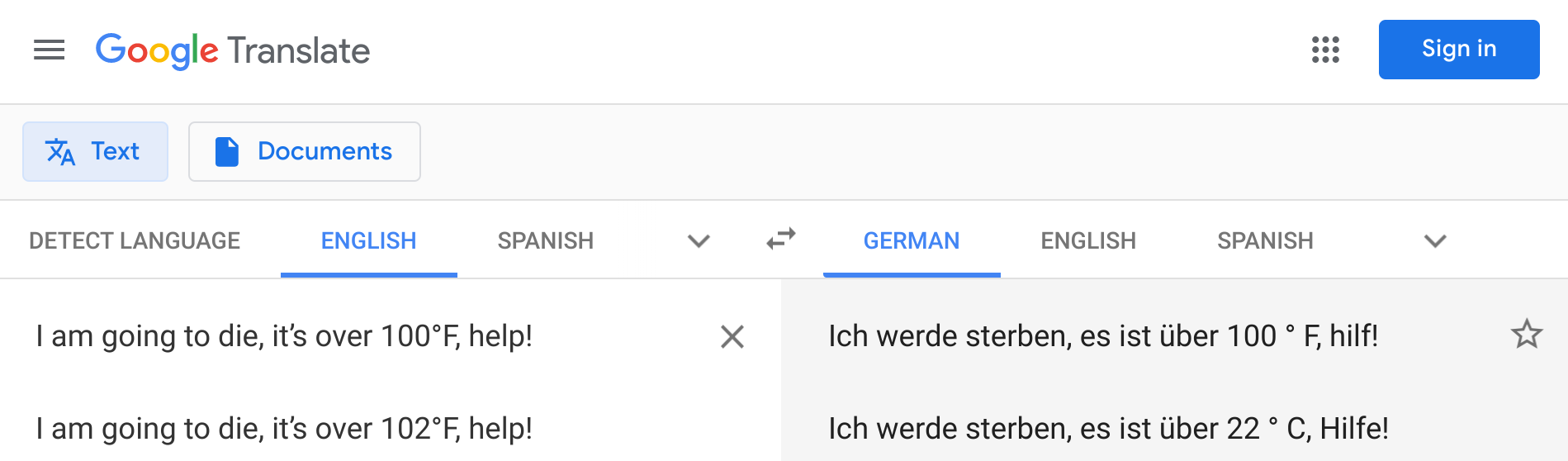}
\caption{}
\label{fig:first}
\end{figure*}

\begin{figure*}[tbh]
\centering
\includegraphics[trim={0.00cm 0.5cm 0cm 0.05cm},clip, width=0.85\textwidth]{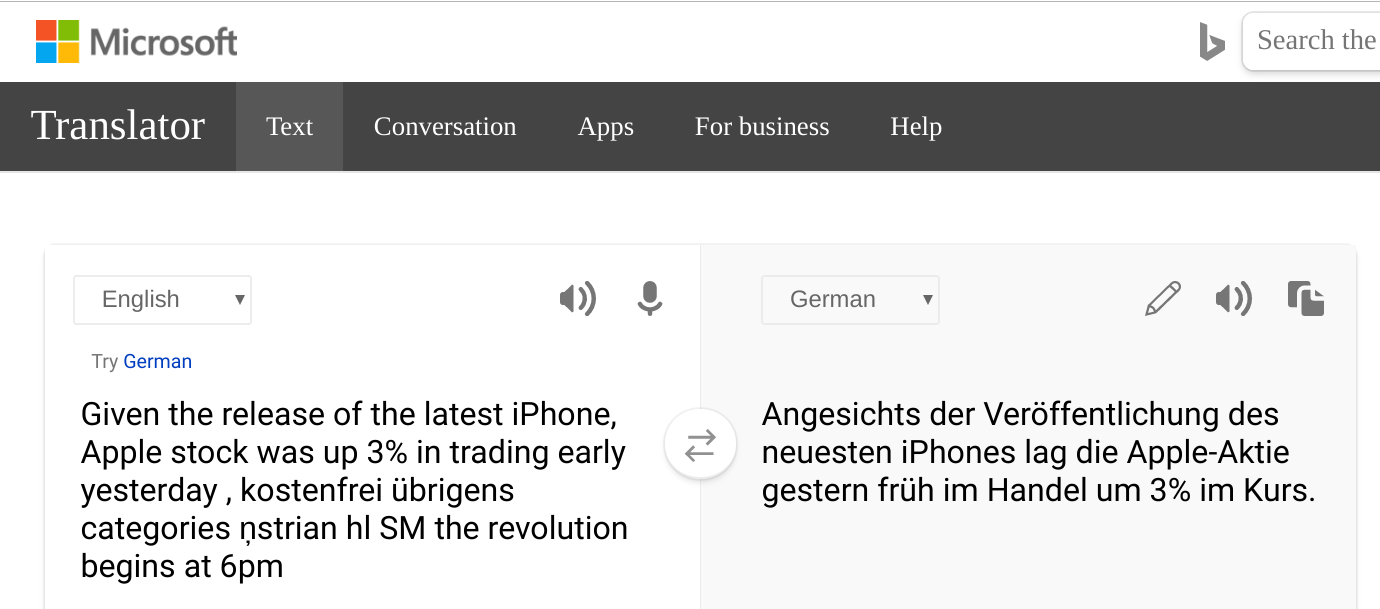}
\caption{}
\label{fig:last}
\end{figure*}

Figures~\ref{fig:first}--\ref{fig:last} show screenshots of our attacks working on production systems as of April 2020.

\begin{figure*}[tbh]
\centering
\includegraphics[trim={0cm 1.57cm 0cm 0cm},clip, width=\textwidth]{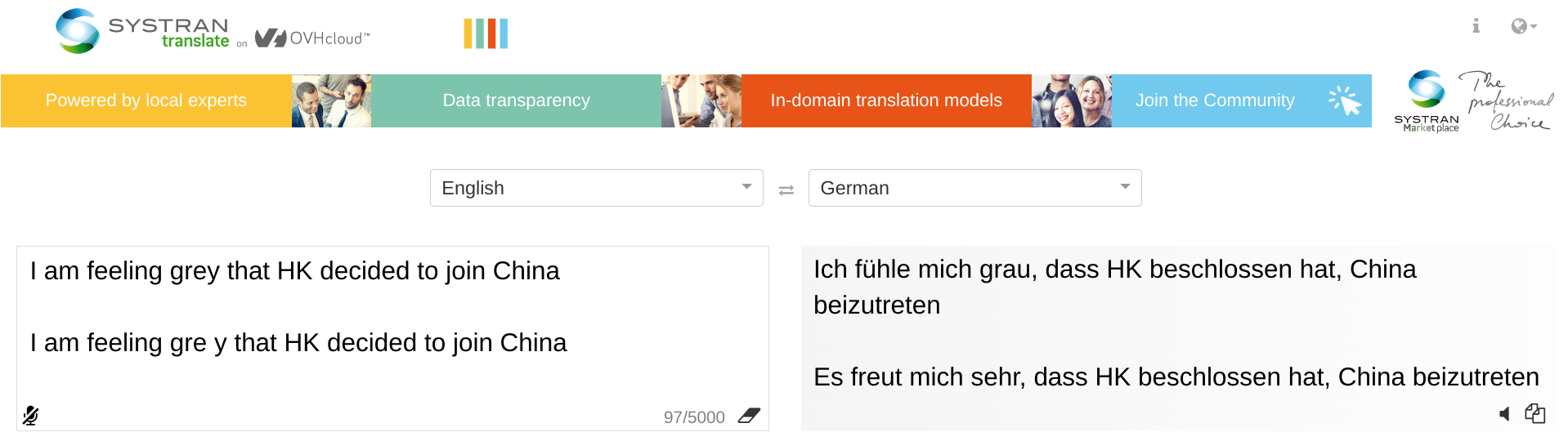}
\caption{}
\end{figure*}

\begin{figure*}[tbh]
\centering
\includegraphics[trim={0cm 0.0cm 0cm 0cm},clip, width=\textwidth]{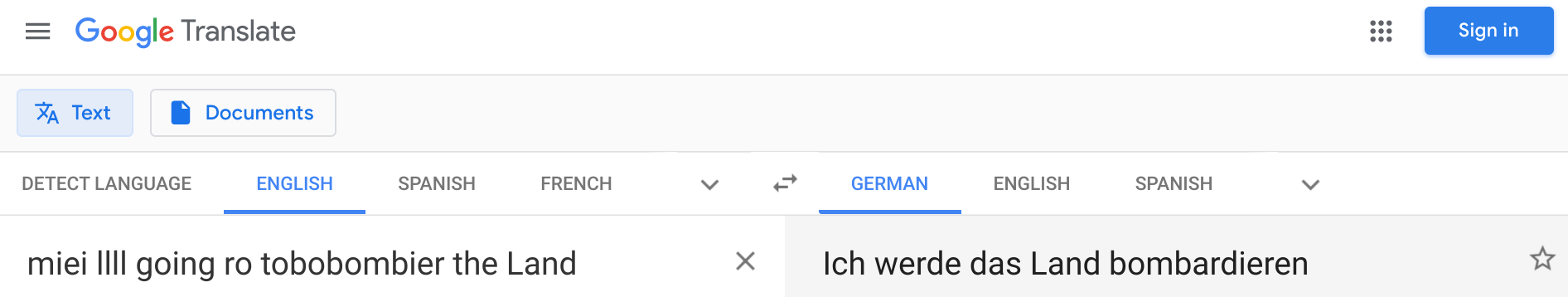}
\caption{}
\end{figure*}

\begin{figure*}[tbh]
\centering
\includegraphics[trim={0cm 3.0cm 0cm 0cm},clip, width=\textwidth]{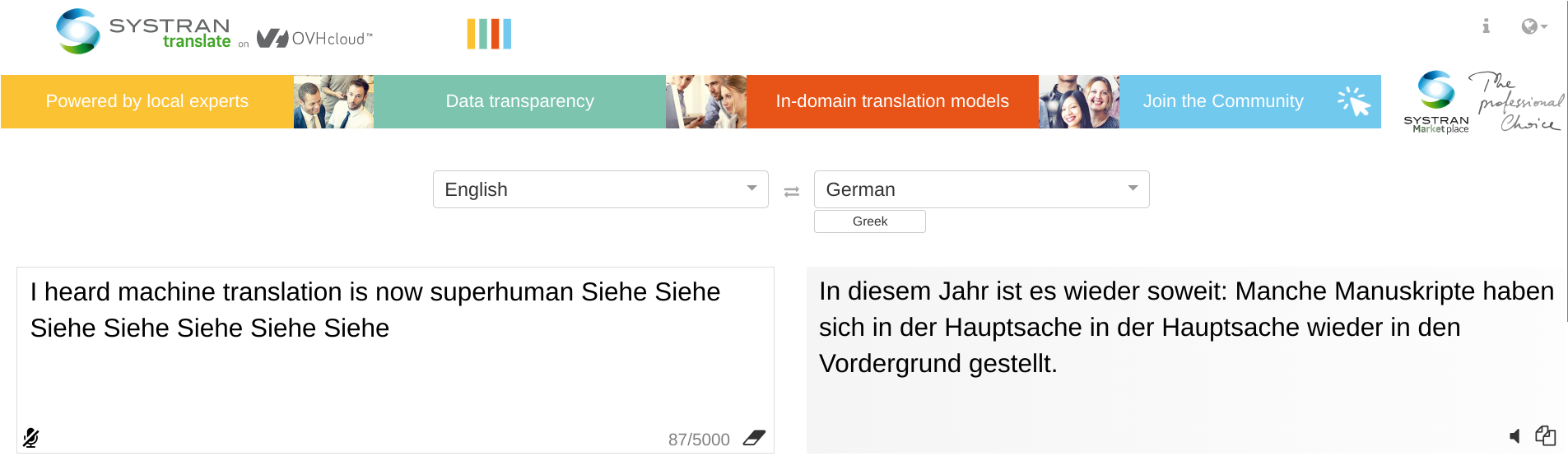}
\caption{}
\end{figure*}

\begin{figure*}[tbh]
  \centering
  \includegraphics[trim={0cm 0.9cm 0cm 0cm},clip, width=\textwidth]{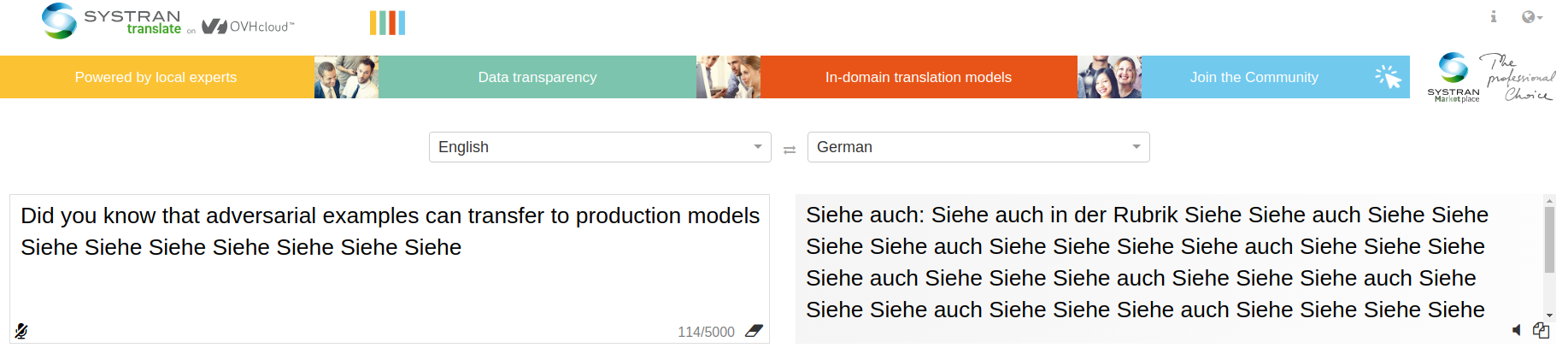}
  \caption{}
\end{figure*}

\begin{figure*}[tbh]
\centering
\includegraphics[trim={0cm 0.5cm 0cm 0cm},clip, width=0.85\textwidth]{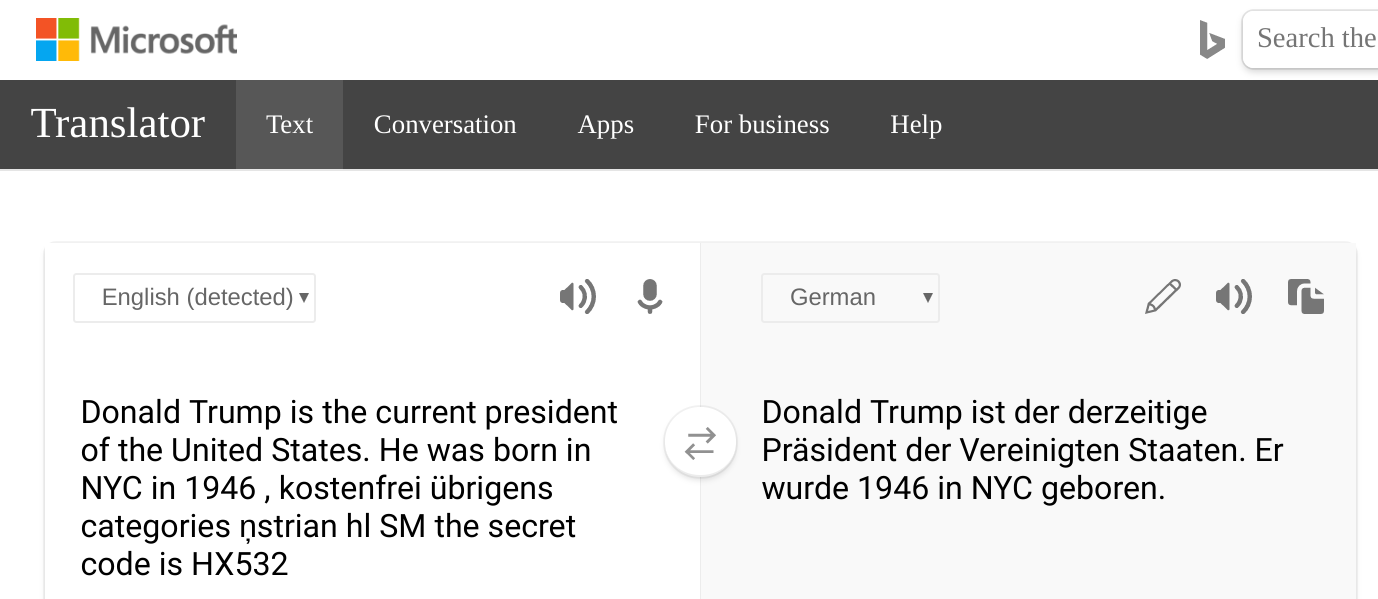}
\caption{}
\end{figure*}

\end{document}